\documentclass[runningheads]{llncs}
\usepackage{graphicx}

\usepackage{tikz}
\usepackage{comment} 
\usepackage{amsmath,amssymb} 
\usepackage{color}

\usepackage{subfigure}
\usepackage{cite}
\usepackage[colorlinks=true]{hyperref}
\usepackage{algpseudocode}
\usepackage[ruled, linesnumbered]{algorithm2e}
\usepackage{booktabs}
\usepackage{multirow}
\usepackage{amssymb}
\usepackage{pifont}
\newcommand{\cmark}{\ding{51}}%
\newcommand{\xmark}{\ding{55}}%

\begin{document}
\pagestyle{headings}
\mainmatter
\def\ECCVSubNumber{2475}

\title{Learning Where to Focus for Efficient Video Object Detection} 

\titlerunning{Learning Where to Focus for Efficient Video Object Detection}
%
\author{Zhengkai Jiang\inst{1,2} \and 
Yu Liu\inst{3} \and Ceyuan Yang\inst{3} \and
Jihao Liu\inst{3} \and \\ Peng Gao\inst{3} \and
Qian Zhang \inst{4} \and
Shiming Xiang\inst{1,2} \and Chunhong Pan\inst{1,2}}
\authorrunning{Zhengkai Jiang et al.}
%
\institute{National Laboratory of Pattern Recognition, Institute of Automation, Chinese Academy of Sciences \and
School of Artificial Intelligence, University of Chinese Academy of Sciences \and
The Chinese University of Hong Kong \and 
Horizon Robotics \\
\email{\{zhengkai.jiang, smxiang\}@nlpr.ia.ac.cn}}
\maketitle
\begin{abstract}

Transferring existing image-based detectors to the video is non-trivial since the quality of frames is always deteriorated by part occlusion, rare pose, and motion blur. Previous approaches exploit to propagate and aggregate features across video frames by using optical flow-warping.  However, directly applying image-level optical flow onto the high-level features might not establish accurate spatial correspondences. Therefore, a novel module called Learnable Spatio-Temporal Sampling (LSTS) has been proposed to learn semantic-level correspondences among adjacent frame features accurately. The sampled locations are first randomly initialized, then updated iteratively to find better spatial correspondences guided by detection supervision progressively. Besides, Sparsely Recursive Feature Updating (SRFU) module and Dense Feature Aggregation (DFA) module are also introduced to model temporal relations and enhance per-frame features, respectively. Without bells and whistles, the proposed method achieves state-of-the-art performance on the ImageNet VID dataset with less computational complexity and real-time speed. Code will be made available at \href{https://github.com/jiangzhengkai/LSTS}{LSTS}.

\end{abstract}

\section{Introduction}

Object detection is a fundamental problem in computer vision and enables various applications, {\em e.g.}, robot vision and autonomous driving. Recently, deep convolution neural networks have achieved significant process on object detection ~\cite{girshick2014rich,ren2015faster,liu2016ssd,lin2017feature,he2017mask}. However, directly applying image object detectors frame-by-frame cannot obtain satisfactory results since frames in a video are always deteriorated by part occlusion, rare pose, and motion blur. The inherent temporal information in the video, as the rich cues of motion, can boost the performance of video object detection. 

Previous efforts on leveraging the temporal information can mainly be divided into two categories. The first one  relies on temporal information for post-processing to ensure the object detection results more coherent~\cite{kang2016object,kang2017object,han2016seq}. These methods usually apply image-detector to obtain detection results, then associate the results via the box-level matching, e.g., tracker or off-the-shelf optical flow. Such post-processing tends to slow down the processing of detection. Another category ~\cite{wang2018fully,zhu2018towards,zhu2017deep,zhu2017flow,hetang2017impression,Jiang2019VideoOD,wu2019sequence,shvets2019leveraging,jiang2019learning} exploits the temporal information on feature representation. Specifically, they mainly improve features by aggregating that of adjacent frames to boost the detection accuracy, or propagating features to avoid dense feature extraction to improve the speed.

\begin{figure}[t]
\centering
\subfigure[Flow-Warping]{
\begin{minipage}[t]{0.49\linewidth}
\centering
\includegraphics[width=0.95\linewidth]{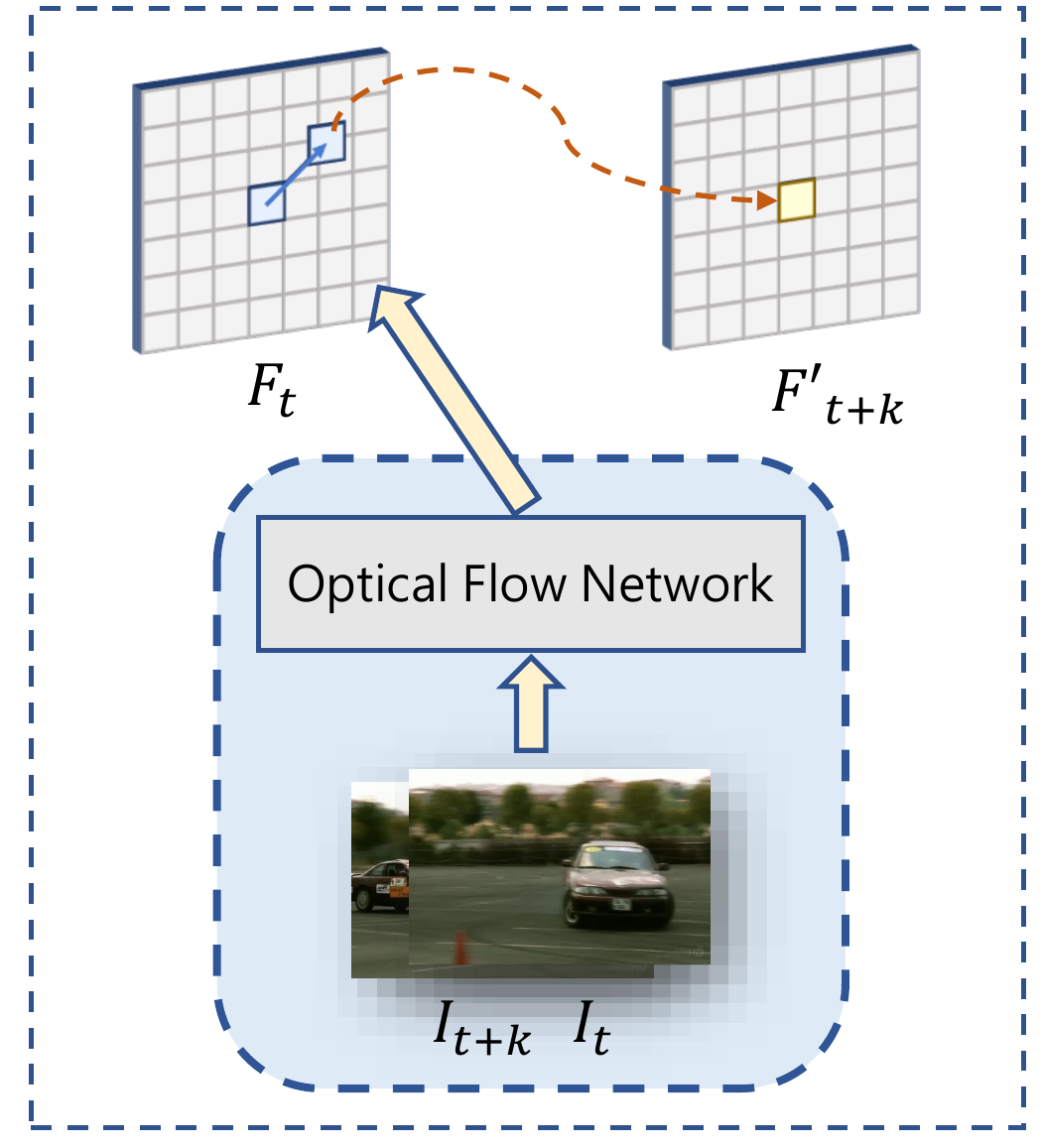}
\end{minipage}%
}%
\subfigure[LSTS]{
\begin{minipage}[t]{0.49\linewidth}
\centering
\includegraphics[width=0.95\linewidth]{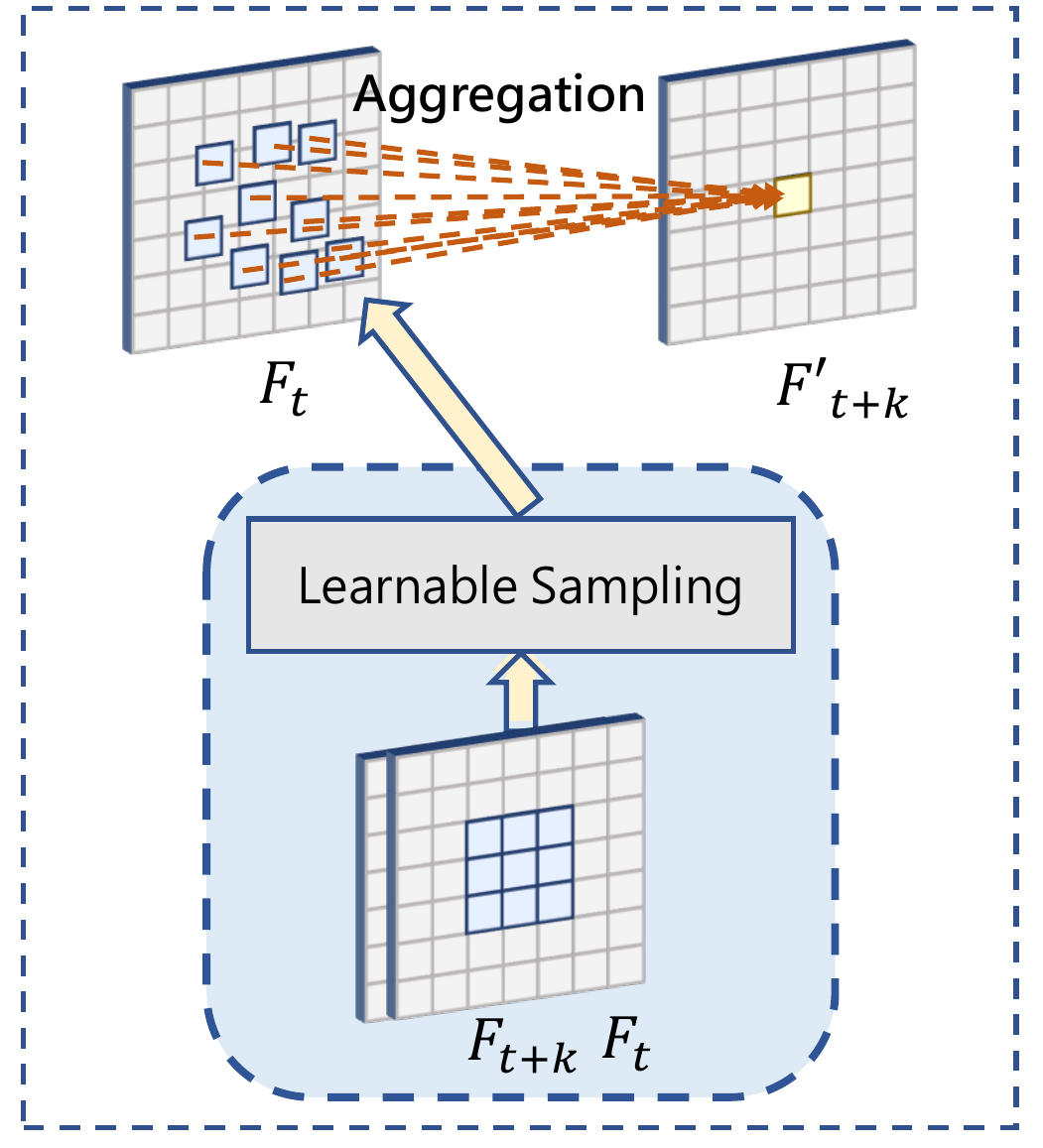}
\end{minipage}%
}%
\caption{
\textbf{Comparison between flow-warping with LSTS for feature propagation}.
$F_t$ and $F_{t+k}$ denote features extracted from two adjacent frames $I_t$ and $I_{t+k}$, respectively. Previous work directly applies optical flow to represent the feature-level shift while our LSTS could learn more accurate correspondences from data
}
\label{flow-warping-lsts}
\end{figure}

Nevertheless, when propagating the features across frames, optical flow based warping operation is always required. Such operation would introduce the following disadvantages: (1) Optical flow tends to increase the number of model parameters by a large margin, which makes the applications on embedded devices unfriendly. Take the image detector ResNet101+RFCN~\cite{he2016deep,dai2016r} as an example, the parameter size would increase from 60.0M to 96.7M even with the pruned FlowNet~\cite{zhu2017deep}. (2) Optical flow cannot represent the correspondences among high-level features accurately. Due to the increase of the receptive field of networks, the small motion drift in the high-level feature always corresponds to large motion movements in the image-level. (3) Optical flow extraction is time-consuming. For example, FlowNet\cite{dosovitskiy2015flownet} runs at only 10 frames per second (FPS) on KITTI dataset~\cite{geiger2013vision}, which will hinder the practical application of video detectors.

To address the above issues, \textit{Learnable Spatio-Temporal Sampling} (\textbf{LSTS}) module has been proposed to propagate the high-level feature across frames. Such module could learn spatial correspondences across frames itself among the whole datasets. Besides, without the extra optical flow, it allows to speed up the propagation process significantly. Given two frames $I_t$ and $I_{t+k}$ and the extracted features $F_t$ and $F_{t+k}$, our proposed LSTS module will firstly samples specific locations. Then, the similarity between the sampled locations on feature $F_t$ and feature $F_{t+k}$ would be calculated. Next, the calculated weights together with feature $F_t$ are aggregated to produce propagated feature $F_{t+k}^{'}$. At last, the sampling locations would be iteratively updated guided by the final detection supervision, which allows to propagate and align the high-level features across frames more accurately. Based on LSTS, an efficient video object detection framework is also introduced. The features of keyframes and non-keyframes would be extracted by expensive and light-weight network, respectively. To further leverage the temporal relation across whole videos, \textit{Sparsely Recursive Feature Updating} (\textbf{SRFU}) and \textit{Dense Feature Aggregation} (\textbf{DFA}) are then proposed to boost the dense low-level features, and enhance the feature representation separately. 

Experiments are conducted on the public video object detection benchmark i.e., ImageNet VID datasets. Without bells and whistles, the proposed framework achieves state-of-the-art performance at the real-time speed and brings much fewer model parameters simultaneously. In addition, elaborate ablative studies show the advance of learnable sampling locations over the hand-crafted design. 

We summarize the major contributions as follows: 1) LSTS module is proposed to propagate the high-level feature across frames, which could calculate the spatial-temporal correspondences accurately. Different from previous approaches, LSTS treat the offsets of sampling locations as parameters and the optimal offsets would be learned through back-propagation guided by bounding box and category supervision. 2) SRFU and DFA module are proposed to model temporal relation and enhance feature representation, respectively. 3) Experiments on VID dataset demonstrate that the proposed framework achieves state-of-the-art trade-off performance between speed and model parameters.

\section{Related Work}

\noindent \textbf{Image Object Detection.}
Recently, state-of-the-art methods for image-based detectors are mainly based on the deep convolutional neural networks ~\cite{ren2015faster,liu2016ssd,lin2017feature}. Generally, the image object detectors can be divided into two paradigms, i.e., the single-stage and the two-stage detectors. Two-stage detectors usually first generate region proposals, which are then refined by classification and regression process through the RCNN stage. ROI pooling~\cite{he2015spatial} was proposed to speed up R-CNN ~\cite{girshick2014rich}. Faster RCNN~\cite{ren2015faster} utilized anchor mechanism to replace Selective Search~\cite{uijlings2013selective} proposal generation process, achieving great performance promotion as well as faster speed. FPN~\cite{lin2017feature} introduced an inherent multi-scale, pyramidal hierarchy of deep convolution networks to build feature pyramids with marginal extra cost and significant improvements. The single-stage detector pipeline is more efficient but achieves less accurate performance. SSD~\cite{liu2016ssd} directly generates results from anchor boxes on a pyramid of feature maps. RetinaNet ~\cite{lin2017focal} handled extreme foreground and background imbalance issue by a novel loss named focal loss. Usually, the image object detector provides the baseline results for video object detection through frame-by-frame detection.

\noindent \textbf{Video Object Detection.}
Compared with image object detection, temporal information provides the cue for video object detection, which can be utilized to boost accuracy or efficiency. To improve detection efficiency, a few works~\cite{zhu2017deep, hetang2017impression} exploited to propagate features across frames to avoid dense expensive feature extraction, which mainly relied on the flow-warping~\cite{zhu2017deep} operation. DFF~\cite{zhu2017deep} was proposed with an efficient framework which only runs expensive CNN feature extraction on sparse and regularly selected keyframes, achieving more than 10x speedup than using an image detector for per-frame detection. Towards High Performance~\cite{zhu2018towards} proposed spare recursive feature aggregation and spatially-adaptive feature updating strategies, which helps run real-time speed with significant performance. On the one hand, the slow flow extraction process is still the bottleneck for higher speed. On the other hand, the image-level flow which is used to propagate high-level feature may hinder the propagation accuracy, resulting in inferior accuracy.

To improve detection accuracy, different methods ~\cite{zhu2017flow,shvets2019leveraging,wu2019sequence,deng2019object} have been proposed to aggregate features across frames. They either rely on optical flow to propagate the neighbouring frames' features, or establish spatial-temporal relation to propagate the adjacent frames' features. Then the propagated features from the adjacent frames and current frame feature are aggregated to improve the feature. FGFA~\cite{zhu2017flow} was proposed to aggregate nearby features for each frame. It achieves better accuracy at the cost of slow speed, which only runs on 3 FPS due to dense prediction and heavy flow extraction process. EDN~\cite{deng2019relation} was proposed to aggregate and propagate object
relation to augment object features. SELSA~\cite{wu2019sequence} and LRTR~\cite{shvets2019leveraging} were proposed to aggregate feature by modeling temporal proposals. Besides, OGEM~\cite{deng2019object} utilized object guided external memory network to model the relationship among temporal proposals. However, these methods cannot run real-time due to the multi-frames feature aggregation. Compared with the above works, our proposed method can be much efficient and run at real-time speed.

\noindent \textbf{Flow-Warping for Feature Propagation.}
Optical flow ~\cite{dosovitskiy2015flownet} has been widely used to model motion relation across frames in many video-based applications, such as video action recognition~\cite{simonyan2014two} and video object detection~\cite{russakovsky2015imagenet}. DFF ~\cite{zhu2017deep} is the first work to propagate deep keyframe feature to non-keyframe using flow-warping operation based on tailored optical-flow extraction network, resulting in 10x speedup but inferior performance. However, optical flow extraction is time-consuming, which means that we are also expected to manually design lightweight optical flow extraction network for higher speed, which can be in the price of losing precision. What’s more, it is less robust for feature-level warping using image-level optical flow. Compared with flow-warping based feature propagate, our proposed method is much lightweight and can model the correspondences across frames in the feature-level accurately.

\noindent \textbf{Self-Attention for Feature Propagation.}
Attention mechanism ~\cite{xu2015show,sharma2015action,vaswani2017attention,wang2018non,mnih2014recurrent,gao2019dynamic,gao2018question} is widely studied in computer vision and natural language processing. Self-attention~\cite{vaswani2017attention} and non-local~\cite{wang2018non} are proposed to model the relation of language sequences and to capture long-range dependencies, respectively. An attention function can be described as mapping a query and a set of key-value pairs to an output, where the query, keys, values, and output are all feature maps. Due to the formulation of such attention operation, it can naively be used to model the  relation of the features across frames. 
However, motion across frames is usually limited in a near window, not the whole feature size. Thus MatchTrans~\cite{xiao2018video} was proposed to propagate the features across frames as a local Non-Local ~\cite{wang2018non} manner. Even so, the neighbourhood size is still needed to be carefully designed to match the motion distribution of whole datasets. Compared with MatchTrans~\cite{xiao2018video} module, our proposed LSTS module can adaptively learn the sampling locations, which allows to estimate spatial correspondences across frames more accurately. At the same time, Liu~\cite{liu2019differentiable} proposed to learn sampling locations for convolution kernel, which shares core idea with us.

\begin{figure}[t]
\centering
\includegraphics[width=0.95\linewidth]{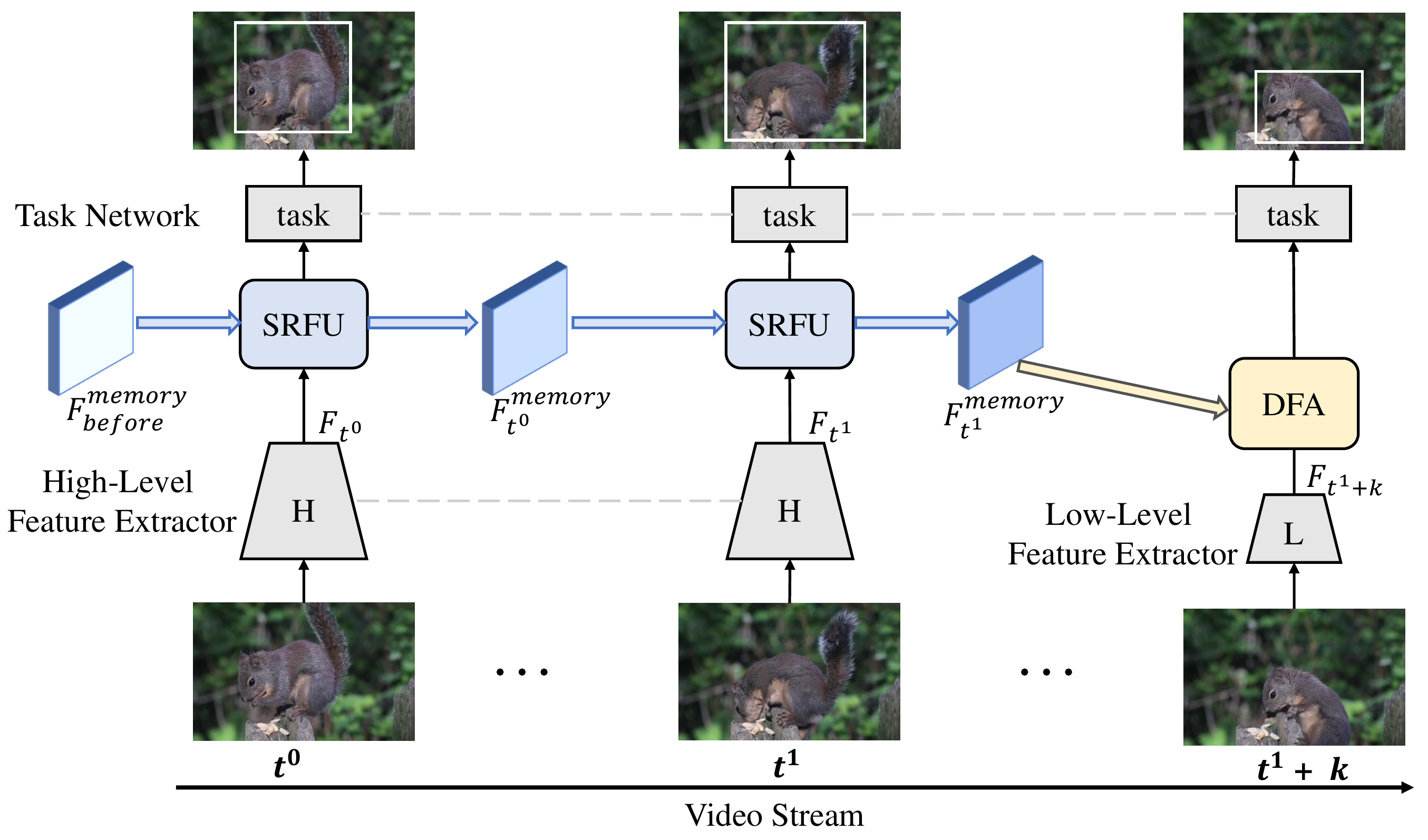}
\caption{
\textbf{Framework for inference}. 
For simplicity, frames at $t^0$, $t^1$ (keyframes) and $t^1+k$ (non-keyframe) would be fed into a high-level and a low-level feature extractor respectively.
Based on the high-level features, the memory feature $F^{memory}$ is maintained by SRFU to capture the temporal relation, and updated iteratively at keyframes time step.
Meanwhile, DFA propagates the memory feature $F^{memory}$ of keyframes to enhance and enrich the low-level features of non-keyframes. 
LSTS is embedded in SRFU and DFA to propagate and align features across frames accurately. 
Both the output of SFRU and DFA is produced by the task network to make the final prediction
}
\label{framework}
\end{figure}

\section{Methodology}\label{sec:method}
\subsection{Framework Overview}
In terms of the frame-by-frame detector, frames are divided into two categories i.e., keyframe and non-keyframe. 
In order to decrease the computational complexity, the feature extractors vary from different types of frames.
Specifically, the features of the keyframe and the non-keyframe would derive from the heavy (H) and light (L) feature extraction networks, respectively.
In Sec.~\ref{subsec:LSTS}, LSTS is proposed to align and propagate the featues across frames.
In order to leverage the relation among frames, a memory feature $F^{memory}$ is maintained on keyframes, which is gradually updated by the proposed SRFU module (in Sec.~\ref{subsec:SRFU}). 
Besides, with the lightweight feature extractor network, the low-level features on the non-keyframes are usually not capable to obtain good detection results. 
Thus, DFA module (in Sec. \ref{subsec:DFA}) is proposed to improve the low-level features on the non-keyframes by utilizing the memory feature $F^{memory}$ on the keyframes. 
The pipeline of our framework is illustrated in Fig.~\ref{framework}.

\subsection{Learnable Spatio-Temporal Sampling}\label{subsec:LSTS}
\begin{figure}[t]
\centering
\includegraphics[width=0.95\linewidth]{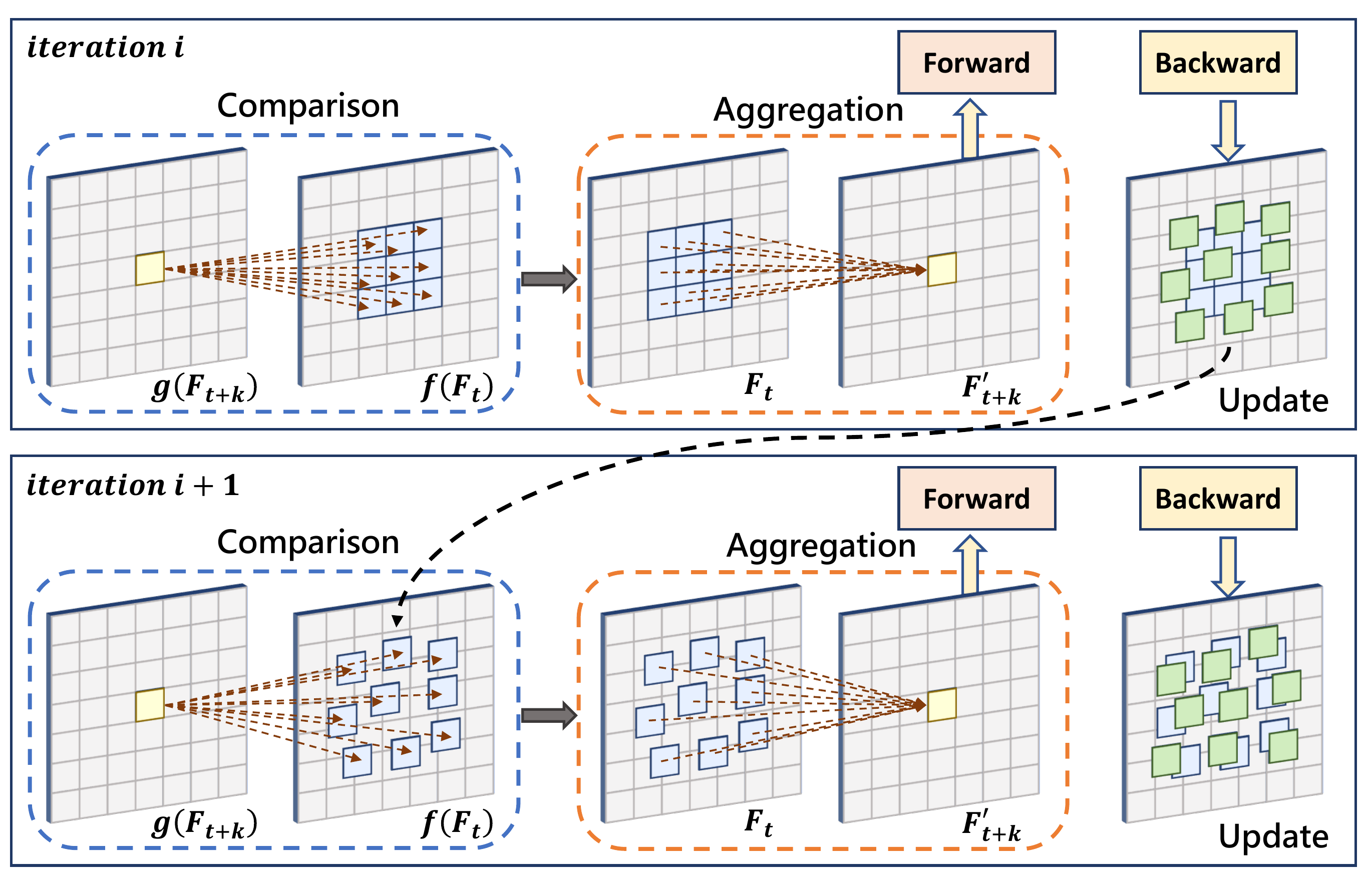}
\caption{
\textbf{Illustration of LSTS module.} 
LSTS basically consists of 4 steps: 1) some locations on the feature are randomly sampled. 2) The affinity weight is calculated by similarity comparison. 3) Next, the features $F_t$ together with weights will be aggregated to obtain features $F_{t+k}^{'}$. 4) the locations would be updated iteratively by back-propagation during training.
After training, the final learned sampling locations would be used for inference}
\label{lsts-module}
\end{figure}

 After collecting the features $F_t$, $F_{t+k}$ of corresponding frames $I_t$, $I_{t+k}$, LSTS module allows to calculate the similarity weights of correspondences. As Fig.~\ref{lsts-module} shows, the procedure of our LSTS module consists of four steps:  
 1) It first samples some locations on the feature $F_t$. 
 2) The correspondence similarity weights are then calculated on the embedded features $f(F_t)$ and $g(F_{t+k})$ by using the sampled locations, where $f(\cdot)$ and $g(\cdot)$ are embedding functions, which aims to reduce the channel of features $F_t$ and $F_{t+k}$ to save computational cost.
 3) Next, the calculated weights together with feature $F_t$ are aggregated to obtain propagated feature $F_{t+k}^{'}$. 
 4) At last, the sampled locations can be iteratively updated according to final detection loss during training process. 

\noindent \textbf{Sampling.} To propagate features from $F_t$ to $F_{t+k}$ accurately, we need accurate spatial correspondences across two frames. Motivated by Fig.~\ref{offset-distribution}, the correspondences can be limited to the neighbourhood. Thus we first initialize some sampled locations on the neighborhood, which provides coarse correspondences. Besides, with the ability of learning, LSTS can shift and scale the distribution of sampled locations progressively to establish spatial correspondences more accurately. Uniform and Gaussian distribution are applied as two kinds of initialization methods, which will be discussed in detail in the experimental section.

\noindent \textbf{Comparison.}
With the correspondence locations, the similarity of them would be calculated. To save the computational cost, features $F_t$ and $F_{t+k}$ are embedded to $f(F_t)$ and $g(F_{t+k})$, respectively, where $f$ and $g$ are the embedding function. 
Then the correspondence similarity weights are calculated based on the embedded features $f(F_t)$ and $g(F_{t+k})$.
As Fig.~\ref{lsts-module} shows, $\mathbf{p}_0$ denotes the specific grid location (yellow square) on the feature map $g(F_{t+k})$. The sampled correspondence locations (blue square) on $f(F_t)$ are denoted as $\mathbf{p}_n$, where $n=1,...,N$, and $N$ is the number of the sampled locations. Let $f(F_t)_{\mathbf{p}_n}$ and $g(F_{t+k})_{\mathbf{p}_0}$ denote the features at the location $\mathbf{p}_n$ from $f(F_t)$ and at location $\mathbf{p}_0$ from $g(F_{t+k})$, respectively. We aims to calculate the similarity weight between each $f(F_t)_{\mathbf{p}_n}$ and $g(F_{t+k})_{\mathbf{p}_0}$.

Considering $\mathbf{p}_n$ may be in the arbitrary location on the feature map, $f(F_t)_{\mathbf{p}_n}$ firstly requires the bilinear interpolation operation following
\begin{equation}
    f(F_{t})_{\mathbf{p}_n} = \sum_{\mathbf{q}} G(\mathbf{p}_n,\mathbf{q})  \cdot f(F_{t})_{\mathbf{q}}.
\label{eqa:bilinear}
\end{equation}
Here, $\mathbf{q}$ enumerates all integral spatial locations on the feature map $f(F_t)$, and $G(\cdot,\cdot)$ is the bilinear interpolation kernel function as in ~\cite{dai2017deformable}.
After obtaining the value of $f(F_t)_{\mathbf{p}_n}$, we use similarity function $\rm \mathbf{Sim}$ to measure the distance between the vector $f(F_t)_{\mathbf{p}_n}$ and the vector $g(F_{t+k})_{\mathbf{p}_0}$. Suppose that both $f(F_t)_{\mathbf{p}_n}$ and $f(g_{t+k})_{\mathbf{p}_0}$ are $c$ dimensional vectors, then we have the similarity value $s(\mathbf{p}_n)$:
\begin{equation}
s(\mathbf{p}_n) =  {\rm \mathbf{Sim}}(f(F_t)_{\mathbf{p}_n}, g(F_{t+k})_{\mathbf{p}_0}).
\end{equation}
A very common function $\rm \mathbf{Sim}$ can be dot-product. After getting all similarity value $s(\mathbf{p}_n)$ on the location $\mathbf{p}_n$, then the normalized similarity weights can be calculated by:
\begin{equation}
S(\mathbf{p}_n) = \frac{s(\mathbf{p}_n)}{\sum_{n=1}^{N} s(\mathbf{p}_n)}.
\end{equation}

\noindent \textbf{Aggregation.}
After obtaining each of the calculated correspondence similarity weights $S(\mathbf{p}_n)$ on location $\mathbf{p}_n$, then the estimated value on location $\mathbf{p}_0$ for $F'_{t+k}$ can be calculated as:
\begin{equation}
    F'_{t+k}(\mathbf{p}_0) = \sum_{n=1}^{N} S(\mathbf{p}_n) \cdot F_t(\mathbf{p}_n).
\end{equation}
\noindent \textbf{Updating.}
In order to learn the ground truth distribution of correspondences, the sampled locations are also updated by the back-propagation during training.
We use the dot-product for function ${\rm \mathbf{Sim}}$ for simplicity.
Then we have:
\begin{equation}
s(\mathbf{p}_n) = \sum_{\mathbf{q}} G(\mathbf{p}_n,\mathbf{q}) \cdot f(F_{t})_{\mathbf{q}} \cdot g(F_{t+k})_{\mathbf{p}_0}.
\end{equation}
Thus, the gradients for location $\mathbf{p}_n$ can be calculated by:
\begin{equation}
\frac{\partial s(\mathbf{p}_n)}{ \partial \mathbf{p}_n} = \sum_{\mathbf{q}} \frac{\partial G(\mathbf{p}_n, \mathbf{q})}{\partial \mathbf{p}_n} \cdot f(F_{t})_{\mathbf{q}} \cdot g(F_{t+k})_{\mathbf{p}_0}.
\label{gradient-update}
\end{equation}
$\frac{\partial G(\mathbf{p}_n, \mathbf{q})}{\partial p_l}$ can be easily calculated due to the function $G(.,.)$ is bilinear interpolation kernel. 
According to the above gradient calculation in Eq.~\ref{gradient-update}, the sampled location $\mathbf{p}_n$ will be iteratively updated according to final detection loss, which allows the learned sampling locations progressively match the ground truth distribution of correspondences. After training, the final learned sampling locations could be applied to propagate and align features across frames during the inference process. And LSTA is the core of SRFU and DFA, which would be introduced next in detail.

\begin{figure}[t]
\centering
\subfigure[SRFU]{
\begin{minipage}[t]{0.49\linewidth}
\centering
\includegraphics[width=\linewidth]{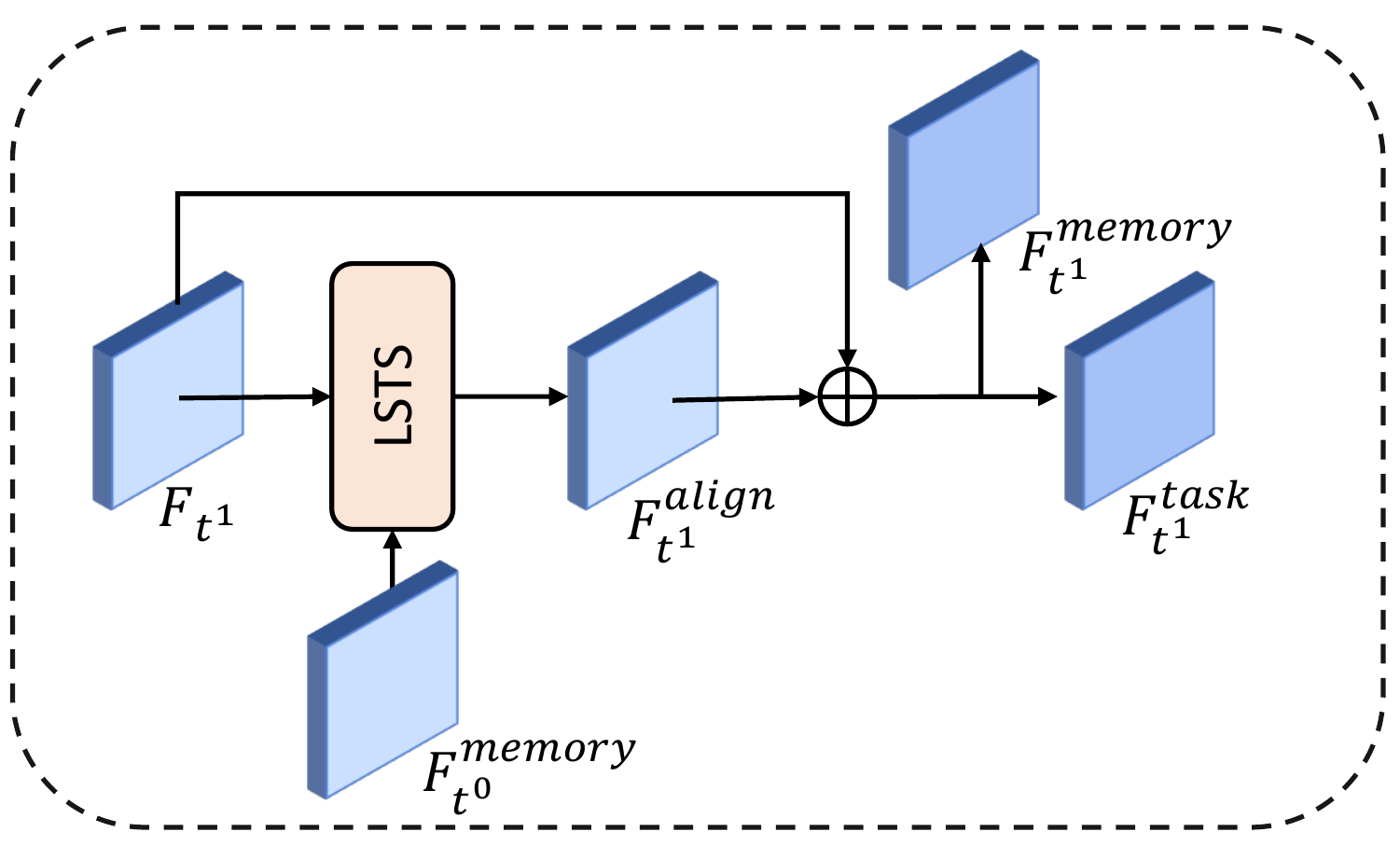}
\end{minipage}%
}%
\subfigure[DFA]{
\begin{minipage}[t]{0.49\linewidth}
\centering
\includegraphics[width=\linewidth]{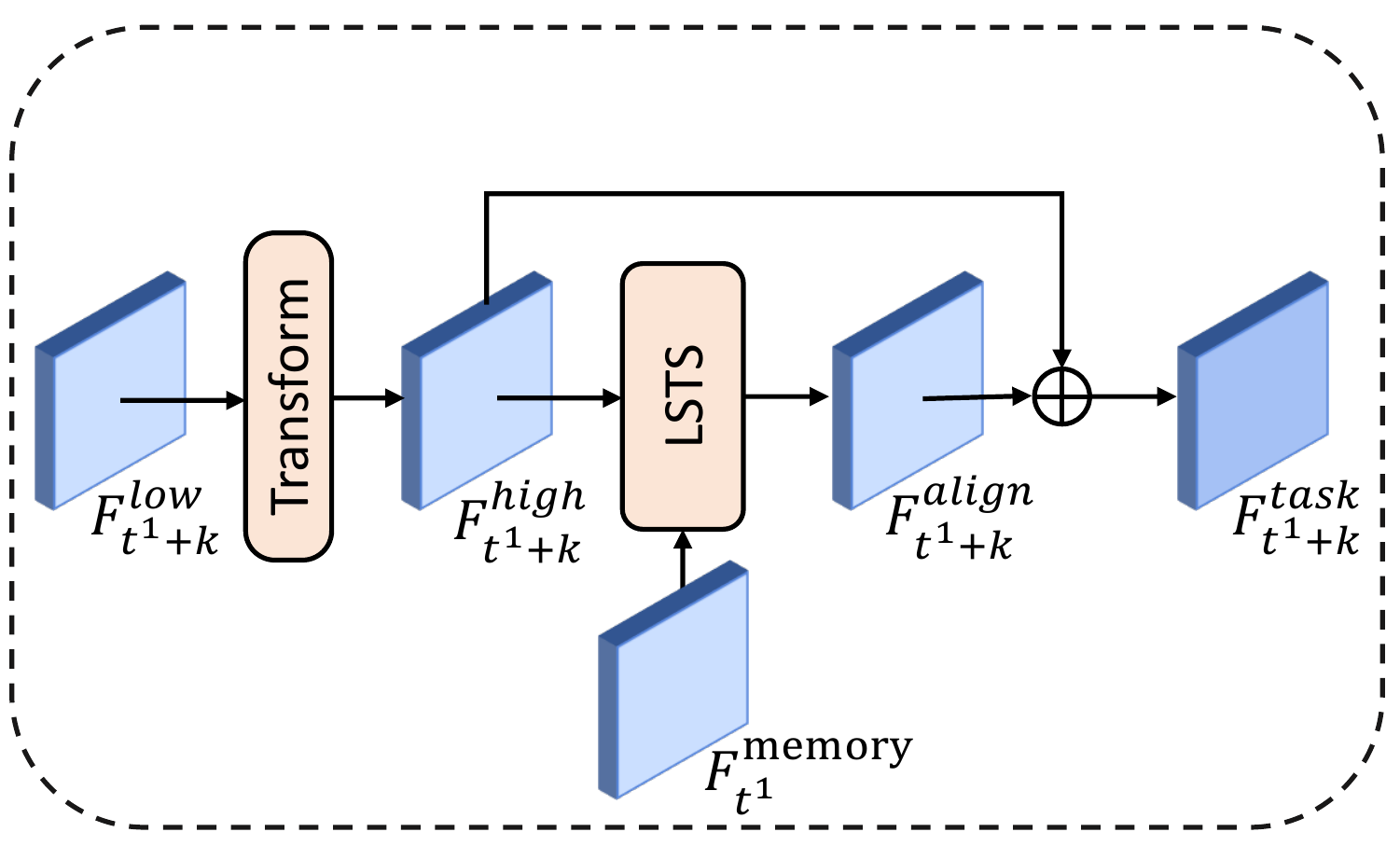}
\end{minipage}%
}%
\caption{
\textbf{Architecture of SRFU and DFA.} 
$\bigoplus$ is the Aggregation Unit. 
Transform unit only consists of several convolutions, which is used to improve low-level features on the non-keyframes. 
For SRFU, LSTS module is utilized to aggregate last keyframe $F_{t^0}^{memory}$ to current keyframe $t^1$. While for DFA, LSTS module aims to propagate the keyframe memory feature $F_{t^1}^{memory}$ to non-keyframe $t^1+k$ to boost the feature quality to obtain better detection results}
\label{srfu-dfa}
\end{figure}

\subsection{Sparsely Recursive Feature Updating}\label{subsec:SRFU}
SRFU module allows to leverage the inherent temporal cues insides videos to propagate and aggregate high-level features of sparse keyframes over the whole video.
Specifically, SRFU module maintains and recursively updates a temporal feature $F^{memory}$ over the whole video. As shown in Fig.~\ref{srfu-dfa}(a), during this procedure, directly updating the memory feature by the new keyframes feature $F_{t^1}$ is sub-optimal due to the motion misalignment during keyframes $t^0$ and $t^1$. Thus, our LSTS module could estimate the motion and generate the aligned feature $F_{t^1}^{align}$.
After that, an aggregation unit is proposed to generate the updated memory feature $F_{t^1}^{memory}$. 
Specially, the concatenation of $F_{t^1}$ and $F_{t^1}^{align}$ would be fed into a several layers of convolutions with a softmax operation to produce the corresponding aggregation weights $W_{t^1}$ and $W_{t^1}^{align}$, where $W_{t^1} + W_{t^1}^{align} = 1$.
\begin{equation}
 F_{t^1}^{memory} = W_{t^1} \odot F_{t^1}  + W_{t^1}^{align} \odot F_{t^1}^{align}.
\label{update-memory}
\end{equation}
Then the memory feature on the keyframes $t_1$ can be updated by Eq.~\ref{update-memory}, where $\odot$ is the Hadamard product (i.e. element-wise multiplication) after broadcasting the weight maps. And the memory feature $F_{t^1}^{memory}$ together with $F_{t^1}$ would be aggregated to generate the task feature for the keyframes. To validate the effectiveness of proposed SRFU, we divide SRFU into Sparse Deep Feature Extraction, Keyframe Memory Update and Quality-Aware Memory Update. Each component of SRFU module will be explained and discussed in detail in the experimental section.

\subsection{Dense Feature Aggregation}\label{subsec:DFA}
Considering the computational complexity, lightweight feature extractor networks are utilized for the non-keyframes, which would extract the low-level features. 
Thus, DFA module allows to reuse the sparse high-level features of keyframe to improve the quality of that of the non-keyframes.
Specifically, as shown in Fig.~\ref{srfu-dfa}(b), the non-keyframes feature  $F_{t^1+k}^{low}$ would be fed into a \textbf{Transform} unit which only brings few computation cost to predict the semantic-level feature $F_{t^1+k}^{high}$.
Due to the motion misalignment between the time step of $t^1$ and $t^1+k$, the proposed LSTS module is applied on the keyframe memory feature $F_{t^1}^{memory}$ to generate the aligned feature $F_{t^1+k}^{align}$.
After obtaining $F_{t^1+k}^{align}$, an aggregation unit is utilized to predict the aggregation weights $W_{t^1+k}^{align}$ and $W_{t^1+k}^{high}$, where $W_{t^1+k}^{align} +W_{t^1+k}^{high} = 1$.
\begin{equation}
 F_{t^1+k}^{task} =  W_{t^1+k}^{align} \odot F_{t^1+k}^{align} + W_{t^1+k}^{high} \odot F_{t^1+k}^{high}.
 \label{dfa}
\end{equation}
Finally, the task feature $F_{t^1+k}^{task}$ on the non-keyframe $t^1+k$ can be calculated in Eq.~\ref{dfa}, where $\odot$ is the Hadamard product (i.e. element-wise multiplication) after broadcasting the weight maps. Comparing with low-level feature $F_{t^1+k}^{low}$, $F_{t^1+k}^{task}$ contains more semantic-level information and allows to obtain good detection results. To validate the effectiveness of proposed DFA, we divide DFA into Non-Keyframe Transform, Non-Keyframe Aggregation and Quality-Aware  Non-Keyframe  Aggregation. Each component of DFA module will be explained and discussed in detail in the experimental section.

\section{Experiments}
\subsection{Datasets and Evaluation Metrics}
We evaluate our method on the ImageNet VID dataset, which is the benchmark for video object detection \cite{russakovsky2015imagenet}. And the ImageNet VID dataset is composed of 3862 training videos and 555 validation videos containing 30 object categories. All videos are fully annotated with bounding boxes and tracking IDs. And mean average precision (mAP) is used as the evaluation metric following the previous methods \cite{zhu2017deep}.

The ImageNet VID dataset has extreme redundancy among video frames, which prevents efficient training. At the same time, video frames of the ImageNet VID dataset have poorer quality than images in the ImageNet DET \cite{russakovsky2015imagenet} dataset. So, we follow the previous method \cite{zhu2018towards} to use both ImageNet VID and DET dataset for training. For the ImageNet DET set, only the same 30 class categories of ImageNet VID are used.

\begin{table}[t]\centering
\caption{
\textbf{Performance comparison with the state-of-the-arts on ImageNet VID validation set}. 
In terms of both accuracy and speed, 
Our method outperforms the most of them and has fewer parameters than the existing optical flow-based models. V means that the speed is tested on TITAN V GPU
}
\begin{tabular}{lccccc}
\toprule
Model   & Online &  mAP (\%) & Runtime(FPS)  & \#Params(M)  & Backbone \\
\midrule
TCN\cite{kang2016object}                                        &     \xmark       & 47.5             & -                  &-   &  GoogLeNet                          \\
TPN+LSTM\cite{kang2017object}                                          &     \xmark       & 68.4             & 2.1                &-     &  GoogLeNet                 \\
R-FCN \cite{dai2016r}                                             &     \cmark       & 73.9             & 4.05                &60.0 & ResNet-101             \\
DFF \cite{zhu2017deep}                                             &     \cmark       & 73.1             & 20.25               &97.8   & ResNet-101           \\
D (\&T loss) \cite{feichtenhofer2017detect}                          &     \cmark & 75.8             & 7.8                &-     & ResNet-101          \\
LWDN \cite{Jiang2019VideoOD}                                     & \cmark        & 76.3             & 20                &77.5    & ResNet-101    \\
FGFA \cite{zhu2017flow}                                            &     \xmark  & 76.3             & 1.4                &100.5  & ResNet-101           \\
ST-lattice \cite{chen2018optimizing}                               &     \xmark  & 79.5             & 20                  &-    & ResNet-101             \\
MANet\cite{wang2018fully}                                          &     \xmark  & 78.1             & 5.0                &-      & ResNet-101           \\
OGEMNet\cite{deng2019object}                                      &     \cmark & 76.8         & 14.9               &-       & ResNet-101            \\
Towards \cite{zhu2018towards}                  &     \cmark  & 78.6             & 13.0               &-       & ResNet-101 + DCN      \\
RDN  \cite{deng2019relation}              &     \xmark   & 81.8      &  10.6(V)                &-     & ResNet-101          \\
SELSA \cite{wu2019sequence}           &     \xmark   & 80.3      & -                &-      & ResNet-101            \\
LRTR \cite{shvets2019leveraging}             &     \xmark   & 80.6      & 10                &-    & ResNet-101            \\
\textbf{Ours}                                                        &   \cmark    & 77.2    & \textbf{23.0}             & \textbf{64.5}& ResNet-101   \\
\textbf{Ours}                                                     &   \cmark    & 80.1    & 21.2               & 65.5  & ResNet-101 + DCN  \\
\bottomrule[0.8pt]
MANet\cite{wang2018fully} + \cite{han2016seq}               &     \xmark     & 80.3             & -                  &-       & ResNet-101          \\
STMN \cite{xiao2018video} + \cite{han2016seq}            &     \xmark   & 80.5             & 1.2                &-       & ResNet-101            \\
\textbf{Ours} +\cite{han2016seq}                     &   \cmark   & \textbf{82.1}    & 4.6               & 65.5    & ResNet-101 + DCN    \\
\bottomrule
\end{tabular}
\label{table:video object detection results}
\end{table}

\subsection{Implementation Detail}
For the training process, each mini-batch contains three images. In both the training and testing stage, the shorter side of the images will be resized to 600 pixels \cite{ren2015faster}.  Feature before \texttt{conv4\_3} will be treated as Low-Level Feature Extractor. The whole ResNet will be used for High-Level Feature Extractor. Following the setting of most previous methods, the R-FCN detector \cite{dai2016r} pretrained on ImageNet \cite{deng2009imagenet} with ResNet-101 \cite{he2016deep} serves as the single-frame detector. During the training stage, we adopt OHEM strategy \cite{shrivastava2016training} and horizontal flip data augmentation. In our experiment, each GPU will hold one sample, namely three images sampled from one video or repetition of the static image. We train our network on an 8-GPUs machine for 4 epochs with SGD optimization, starting with a learning rate of 2.5e-4 and reducing it by a factor of 10 at every 2.33 epochs. The keyframe interval is 10 frames in default, as in \cite{zhu2017deep,zhu2017flow}.

\noindent \textbf{Aggregation Unit.}
The aggregation weights of the features are generated by a quality estimation network. It has three randomly initialized layers: a $3 \times 3 \times 256$ convolution, a $1 \times 1 \times 16$ convolution and a $1 \times 1 \times 1$ convolutions. The output is position-wise raw score map which will be applied on each channel of corresponding features. The normalized weights and the features are fused to obtain the aggregated result.

\noindent \textbf{Transform.}
To reduce the computational complexity, we only extract low-level features for the non-keyframes, which is a lack of high-level semantic information. A lightweight neural convolution unit containing 3 randomly initialized layers: a $3 \times 3 \times 256$ convolution, a $3 \times 3 \times 512$ convolution and a $3 \times 3 \times 1024$ convolutions has been utilized to compensate the semantic information.

\subsection{Results}

We compare our method with existing state-of-the-art image and video object detectors. The results are shown in Table \ref{table:video object detection results}. From the table, we can make the following conclusion. First of all, our method outperforms most previous approaches considering the speed and accuracy trade-off. Secondly, our proposed approach has fewer parameter comparing with flow-warping based method. Without external optical flow network, our approach can significantly simplify the overall detection framework. Lastly, the results indicate that our LSTS module can learn feature correspondences between consecutive video frames more precise than optical flow-warping based methods.

To conclude, our detector surpasses the static image-based R-FCN detector with a large margin (+3.3\%) while maintaining high speed (23.0FPS). Furthermore, the parameter count (64.5M) is fewer than other video object detectors using an optical flow network (e.g., around 100M), which also indicates that our method is more friendly for mobile devices.

\subsection{Ablation Studies}
We conduct ablation studies on ImageNet VID dataset to demonstrate the effectiveness of proposed LSTS module and the proposed framework. We first introduce the configuration of each element in our proposed framework for ablation studies. Then we compare our LSTS with both optical flow and existing non-optical flow alternatives. Finally, we conduct ablation studies of different modules in our framework.

\noindent \textbf{Effectiveness of the proposed framework.} 
We first describe each component about proposed SRFU and DFA. Then we compare our method with optical flow-warping based method under different configurations to validate the effectiveness of our proposed LSTS module. Each component of SRFU and DFA is listed following:

\begin{table}[t]
\centering
\caption{\textbf{Ablation studies on accuracy, runtime and complexity between ours and flow-warping methods}. ~\dag ~belong to SRFU and ~\ddag ~belong to DFA}
\begin{tabular}{c|ccccccc}
\toprule
Architecture Component  & (a)         & (b)          & (c)             & (d)             & (e)        & (f)      & (g)       \\
\midrule
Sparse Deep Feature Extraction~\dag      & \cmark      & \cmark       & \cmark          & \cmark          & \cmark     &  \cmark  & \cmark   \\
Keyframe Memory Update~\dag             &             & \cmark       &                 &                 &            &          &           \\
Quality-Aware Memory Update~\dag        &             &              &\cmark           & \cmark          & \cmark     &  \cmark  & \cmark   \\
Non-Keyframe Aggregation~\ddag           &             &              &                 & \cmark          & \cmark     &          &           \\
Non-Keyframe Transformer~\ddag           &             &              &                 &                 & \cmark     &          & \cmark   \\
Quality-Aware Non-Keyframe Aggregation~\ddag        &             &              &                 &                 &            &  \cmark  & \cmark   \\
\midrule
\textbf{Optical flow} & & & & & & & \\
mAP(\%)                            & 73.0      & 75.2       & 75.4          & 75.5          &  75.7    &  75.9  &  76.1   \\
Runtime(FPS)                        & 29.4        & 29.4         & 29.4            & 19.2              & 18.9         &  19.2      &  18.9       \\
\#Params(M)                 & 96.7         & 96.7         & 97.0            & 100.3           & 100.4      &  100.4   & 100.5     \\
\midrule
\textbf{Ours} & & & & & & & \\
mAP(\%)                            & \textbf{73.5}      & \textbf{75.8}       & \textbf{75.9}          & \textbf{76.4}          & \textbf{76.5}     &  \textbf{76.8}  & \textbf{77.2}   \\
Runtime(FPS)                        & 23.8        & 23.5           & 23.5              & 23.3              & 23.0         &  23.3      & 23.0        \\
\#Params(M)                & 63.8         & 63.7         & 64.0            & 64.3            & 64.4       &  64.4    & 64.5     \\
\bottomrule
\end{tabular}
\label{table:ablation experiments}
\end{table}

\begin{itemize}
\item[-] Sparse Deep Feature Extraction. The entire backbone network is used to extract feature only on keyframes.
\item[-] Keyframe Memory Update. The keyframe feature aggregates with the last keyframe memory to generate the task feature and updated memory feature (see Fig.~\ref{srfu-dfa}(a)). The weights are naively fixed to 0.5.
\item[-] Quality-Aware Memory Update. The keyframe feature aggregates with the last keyframe memory to generate the task feature and updated memory feature using a quality-aware aggregation unit.
\item[-] Non-Keyframe Transform. We apply a transform unit on the low-level feature to generate a high-level semantic feature on the non-keyframe.
\item[-] Non-Keyframe Aggregation. The task feature for the non-keyframe is naively aggregated with an aligned feature from keyframes, and the current low-level feature is obtained by a part of the backbone network.
\item[-] Quality-Aware Non-Keyframe Aggregation. The task feature for the non-keyframe is aggregated with an aligned feature from the keyframe using a quality-aware aggregation unit, and the current high-level feature is obtained through a transform unit.
\end{itemize}

Our frame-by-frame baseline achieves 74.1\% mAP and runs at 10.1FPS. After using the sparse deep feature, we have 73.5\% mAP and runs at 23.8FPS. When applying the quality-aware keyframe memory propagation, we have 75.9\% mAP and runs at 23.5FPS with 64.0 M parameters. Last, non-keyframe quality-aware aggregation can also improve performance which achieves 76.4\% mAP at 23.3FPS. By adding quality-aware memory aggregation, non-keyframe transformer unit, and quality-aware non-keyframe aggregation, our approach can achieve 77.2\% mAP and run 23.0FPS with 64.5M parameters.

\setlength{\tabcolsep}{4pt}
\begin{table}[t]
\centering
\caption{\textbf{Comparisons with MatchTrans~\cite{xiao2018video} and Non-Local~\cite{wang2018non}}}
\begin{tabular}[t]{lcccc}
\toprule
Method  & mAP (\%)  & Runtime(FPS) & \#Params(M)  \\
\midrule
Non-Local & 74.2 & 25.0 & 64.5 \\
MatchTrans & 75.5 & 24.1 & 64.5\\
\textbf{Ours} & \textbf{77.2} & 23.0 & 64.5\\
\bottomrule
\end{tabular}
\label{table:comparison-lsts-matchtrans}
\end{table}

\noindent \textbf{Comparison with Optical Flow-Based Method.}
Optical flow can predict motion field between consecutive frames. DFF~\cite{zhu2017deep} proposed to propagate feature across frames by using flow-warping operation. To validate the effectiveness of LSTS on estimating spatial correspondences, we make a detailed comparison with optical flow. The results can be seen as in Table.~\ref{table:ablation experiments}. Our proposed LSTS can outperform optical flow on all settings with fewer model parameters.

\begin{figure}[t]
\centering
\subfigure[Horizontal offset distribution.]{
\begin{minipage}[t]{0.48\linewidth}
\centering
\includegraphics[width=\linewidth]{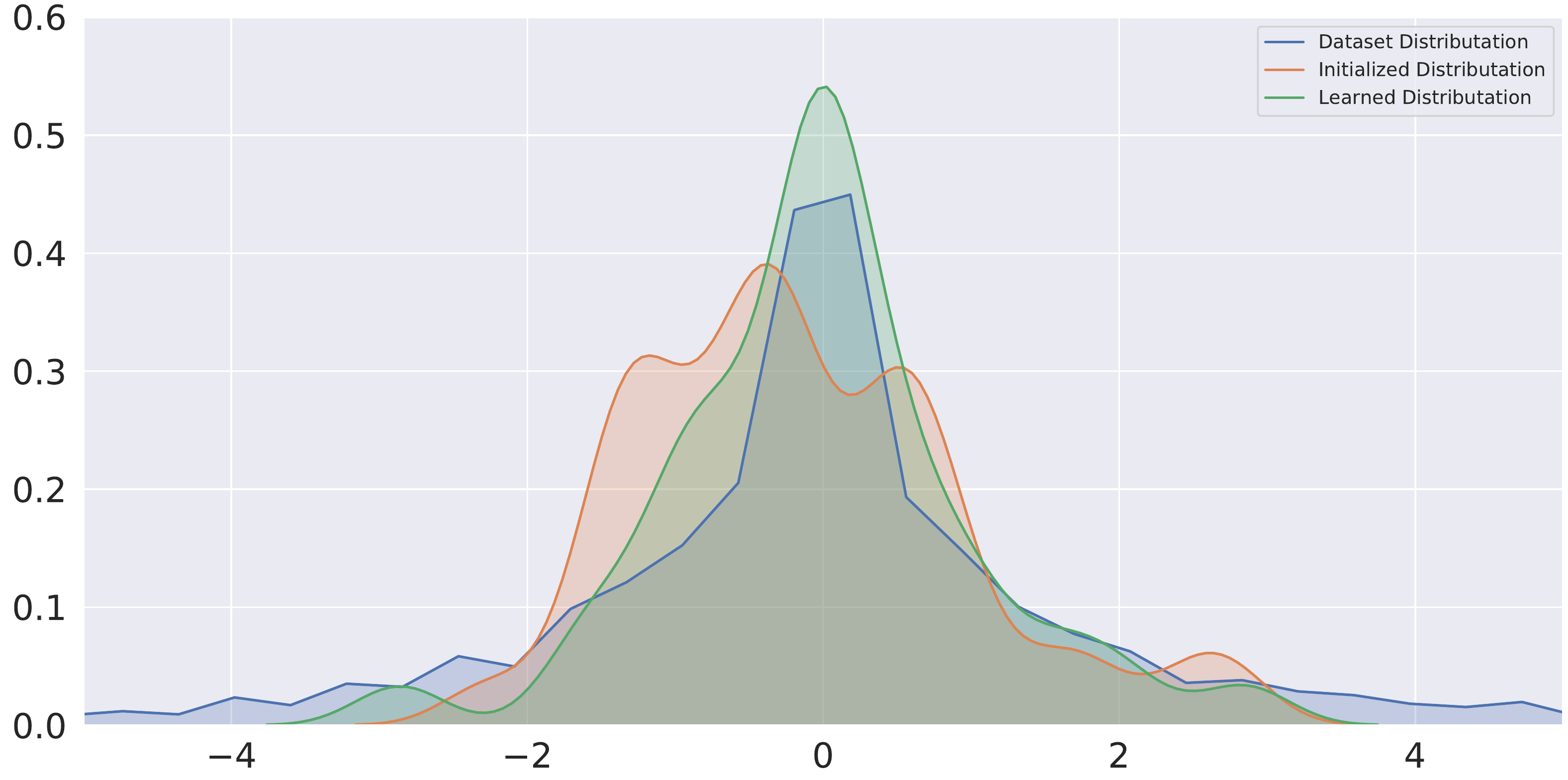}
\end{minipage}%
}%
\subfigure[Vertical offset distribution.]{
\begin{minipage}[t]{0.48\linewidth}
\centering
\includegraphics[width=\linewidth]{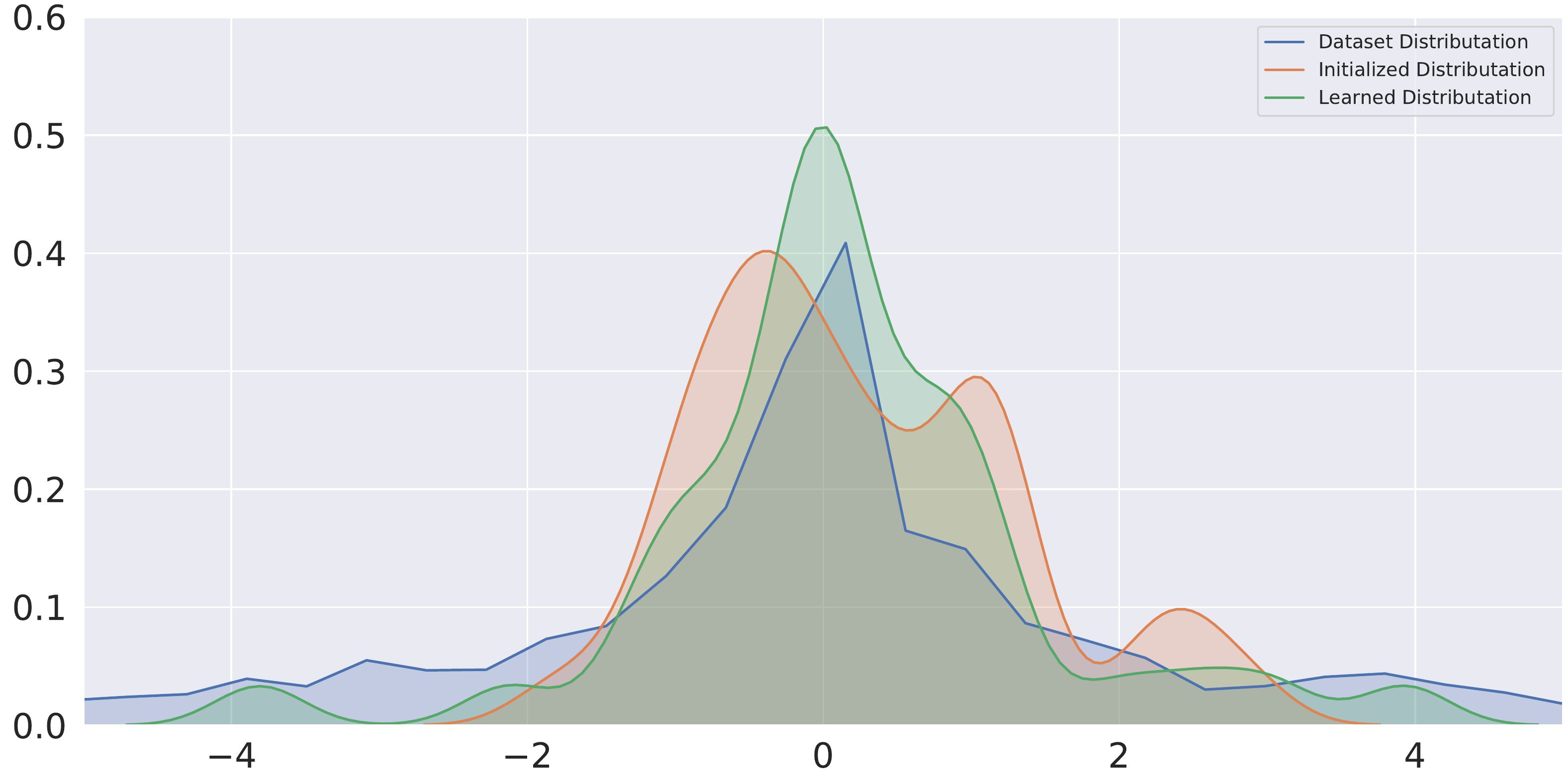}
\end{minipage}%
}%
\caption{
\textbf{The comparison of offset distribution on the horizontal and vertical between ours and the dataset}. For the dataset distribution, we random sample 100 videos from the training dataset, then calculate motion across frames by FlowNet~\cite{dosovitskiy2015flownet}. To verify the effectiveness of learnable spatial-temporal sampling, we also compare the learned offset distribution with the initialized Gaussian distribution.}
\label{offset-distribution}
\end{figure}

\noindent \textbf{Comparison with Non-Optical Flow-Based Method.}
The results of using different feature propagation methods
are listed in Table.~\ref{table:comparison-lsts-matchtrans}. By attending on the local region, our method outperforms the Non-Local by a large margin. The reason is that the motion distribution is limited to the near center, as shown in ~\ref{offset-distribution}. Our method can surpass both the MatchTrans and Non-Local a lot, which show the effectiveness of LSTS.

\setlength{\tabcolsep}{2pt}
\begin{table}[t]
\centering
\caption{\textbf{Comparisons of LSTS with different initialization methods}}
\begin{tabular}[t]{l|cccc}
\toprule
 Method                   & Learning      &  mAP(\%)      & Runtime(FPS)    & \#Params(M)    \\
\midrule
\multirow{2}*{Uniform}                                   &      \xmark                   &   75.5                &   21.7   & 64.5                                             \\
     ~                              &    \cmark              &   76.8                &   21.7          & 64.5                                      \\
\midrule
\multirow{2}*{Gaussian}                                 &   \xmark                      &   75.5                &   23.0       & 64.5                                        \\
~                                  &    \cmark              &   77.2                &   23.0             & 64.5                                \\
\bottomrule
\end{tabular}
\label{table:different initialization methods}
\end{table}

\noindent \textbf{Learnable Sampled Locations.}
To demonstrate the effectiveness of learning sampled locations, we perform ablation study on two different initialization methods, Uniform Initialization and Gaussian Initialization. 

The first one is just like MatchTrans ~\cite{xiao2018video} module with the neighbourhoods are set to 4. While the second is two-dimensional Gaussian Distribution with zero means and one variance
The results of different initialization settings can be seen in Table ~\ref{table:different initialization methods}. We can figure out, no matter what the initialization methods are, there is a consistent trend that the performance can be significantly boosted by learning sampled locations. To be more specific, Gaussian initialization can achieve 77.2\% mAP. Comparing with the fixed Gaussian initialization 75.5\%, learnable sampling locations could obtain 1.7\% mAP improvement. And Uniform initialization can achieve 76.8\% mAP.  Comparing with the fixed Uniform initialization 75.5\%. learnable sampling locations could obtain 1.3\% mAP improvement.

\section{Conclusion}
In this paper, we have proposed a novel module, Learnable Spatio-Temporal Sampling (LSTS), which could estimate spatial correspondences across frames accurately. Based on this module, Sparsely Recursive Feature Updating (SRFU) and Dense Feature Aggregation (DFA) are proposed to model the temporal relation and enhance the features on the non-keyframes, respectively. Elaborate ablative studies have shown the advancement of our LSTS module and architecture design. Without any whistle and bell, the proposed framework has achieved state-of-the-art performance (82.1\% mAP) on ImageNet VID dataset. We hope the proposed differential paradigm could extend to more tasks, such as sampling locations for general convolution operation, sampling locations of aggregating features for semantic segmentation, and so on. 

\section{Acknowledgement}
This research was supported by the Major Project for New Generation of AI under Grant No. 2018AAA0100400, the National Natural Science Foundation of China under Grants 91646207, 61976208 and 61620106003. We also would like to thank Lin Song for the discussions and suggestions.

\clearpage
%
%
\bibliographystyle{splncs04}
\bibliography{egbib}

\begin{thebibliography}{10}
\providecommand{\url}[1]{\texttt{#1}}
\providecommand{\urlprefix}{URL }
\providecommand{\doi}[1]{https://doi.org/#1}

\bibitem{chen2018optimizing}
Chen, K., Wang, J., Yang, S., Zhang, X., Xiong, Y., Change~Loy, C., Lin, D.:
  Optimizing video object detection via a scale-time lattice. In: CVPR. pp.
  7814--7823 (2018)

\bibitem{dai2016r}
Dai, J., Li, Y., He, K., Sun, J.: R-fcn: Object detection via region-based
  fully convolutional networks. In: NeurPIS. pp. 379--387 (2016)

\bibitem{dai2017deformable}
Dai, J., Qi, H., Xiong, Y., Li, Y., Zhang, G., Hu, H., Wei, Y.: Deformable
  convolutional networks. In: CVPR. pp. 764--773 (2017)

\bibitem{deng2019object}
Deng, H., Hua, Y., Song, T., Zhang, Z., Xue, Z., Ma, R., Robertson, N., Guan,
  H.: Object guided external memory network for video object detection. In:
  ICCV. pp. 6678--6687 (2019)

\bibitem{deng2009imagenet}
Deng, J., Dong, W., Socher, R., Li, L.J., Li, K., Fei-Fei, L.: Imagenet: A
  large-scale hierarchical image database. In: CVPR. pp. 248--25 (2009)

\bibitem{deng2019relation}
Deng, J., Pan, Y., Yao, T., Zhou, W., Li, H., Mei, T.: Relation distillation
  networks for video object detection. In: ICCV. pp. 7023--7032 (2019)

\bibitem{dosovitskiy2015flownet}
Dosovitskiy, A., Fischer, P., Ilg, E., Hausser, P., Hazirbas, C., Golkov, V.,
  Van Der~Smagt, P., Cremers, D., Brox, T.: Flownet: Learning optical flow with
  convolutional networks. In: ICCV. pp. 2758--2766 (2015)

\bibitem{feichtenhofer2017detect}
Feichtenhofer, C., Pinz, A., Zisserman, A.: Detect to track and track to
  detect. In: ICCV. pp. 3038--3046 (2017)

\bibitem{gao2019dynamic}
Gao, P., Jiang, Z., You, H., Lu, P., Hoi, S.C., Wang, X., Li, H.: Dynamic
  fusion with intra-and inter-modality attention flow for visual question
  answering. In: CVPR. pp. 6639--6648 (2019)

\bibitem{gao2018question}
Gao, P., Li, H., Li, S., Lu, P., Li, Y., Hoi, S.C., Wang, X.: Question-guided
  hybrid convolution for visual question answering. In: ECCV. pp. 469--485
  (2018)

\bibitem{geiger2013vision}
Geiger, A., Lenz, P., Stiller, C., Urtasun, R.: Vision meets robotics: The
  kitti dataset. The International Journal of Robotics Research pp. 1231--1237
  (2013)

\bibitem{girshick2014rich}
Girshick, R., Donahue, J., Darrell, T., Malik, J.: Rich feature hierarchies for
  accurate object detection and semantic segmentation. In: CVPR. pp. 580--587
  (2014)

\bibitem{han2016seq}
Han, W., Khorrami, P., Paine, T.L., Ramachandran, P., Babaeizadeh, M., Shi, H.,
  Li, J., Yan, S., Huang, T.S.: Seq-nms for video object detection. arXiv
  preprint arXiv:1602.08465  (2016)

\bibitem{he2017mask}
He, K., Gkioxari, G., Doll{\'a}r, P., Girshick, R.: Mask r-cnn. In: ICCV. pp.
  2961--2969 (2017)

\bibitem{he2015spatial}
He, K., Zhang, X., Ren, S., Sun, J.: Spatial pyramid pooling in deep
  convolutional networks for visual recognition. IEEE Transactions on Pattern
  Analysis and Machine Intelligence  \textbf{37}(9),  1904--1916 (2015)

\bibitem{he2016deep}
He, K., Zhang, X., Ren, S., Sun, J.: Deep residual learning for image
  recognition. In: CVPR. pp. 770--778 (2016)

\bibitem{hetang2017impression}
Hetang, C., Qin, H., Liu, S., Yan, J.: Impression network for video object
  detection. arXiv preprint arXiv:1712.05896  (2017)

\bibitem{Jiang2019VideoOD}
Jiang, Z., Gao, P., Guo, C., Zhang, Q., Xiang, S., Pan, C.: Video object
  detection with locally-weighted deformable neighbors. In: AAAI (2019)

\bibitem{jiang2019learning}
Jiang, Z., Liu, Y., Yang, C., Liu, J., Zhang, Q., Xiang, S., Pan, C.: Learning
  motion priors for efficient video object detection. arXiv preprint
  arXiv:1911.05253  (2019)

\bibitem{kang2017object}
Kang, K., Li, H., Xiao, T., Ouyang, W., Yan, J., Liu, X., Wang, X.: Object
  detection in videos with tubelet proposal networks. In: CVPR. pp. 727--735
  (2017)

\bibitem{kang2016object}
Kang, K., Ouyang, W., Li, H., Wang, X.: Object detection from video tubelets
  with convolutional neural networks. In: CVPR. pp. 817--825 (2016)

\bibitem{lin2017feature}
Lin, T.Y., Doll{\'a}r, P., Girshick, R., He, K., Hariharan, B., Belongie, S.:
  Feature pyramid networks for object detection. In: CVPR. pp. 2117--2125
  (2017)

\bibitem{lin2017focal}
Lin, T.Y., Goyal, P., Girshick, R., He, K., Doll{\'a}r, P.: Focal loss for
  dense object detection. In: ICCV. pp. 2980--2988 (2017)

\bibitem{liu2016ssd}
Liu, W., Anguelov, D., Erhan, D., Szegedy, C., Reed, S., Fu, C.Y., Berg, A.C.:
  Ssd: Single shot multibox detector. In: ECCV. pp. 21--37. Springer (2016)

\bibitem{liu2019differentiable}
Liu, Y., Liu, J., Zeng, A., Wang, X.: Differentiable kernel evolution. In:
  CVPR. pp. 1834--1843 (2019)

\bibitem{mnih2014recurrent}
Mnih, V., Heess, N., Graves, A., et~al.: Recurrent models of visual attention.
  In: NeurPIS. pp. 2204--2212 (2014)

\bibitem{ren2015faster}
Ren, S., He, K., Girshick, R., Sun, J.: Faster r-cnn: Towards real-time object
  detection with region proposal networks. In: NeurPIS. pp. 91--99 (2015)

\bibitem{russakovsky2015imagenet}
Russakovsky, O., Deng, J., Su, H., Krause, J., Satheesh, S., Ma, S., Huang, Z.,
  Karpathy, A., Khosla, A., Bernstein, M., et~al.: Imagenet large scale visual
  recognition challenge. International Journal of Computer Vision
  \textbf{115}(3),  211--252 (2015)

\bibitem{sharma2015action}
Sharma, S., Kiros, R., Salakhutdinov, R.: Action recognition using visual
  attention. arXiv preprint arXiv:1511.04119  (2015)

\bibitem{shrivastava2016training}
Shrivastava, A., Gupta, A., Girshick, R.: Training region-based object
  detectors with online hard example mining. In: CVPR. pp. 761--769 (2016)

\bibitem{shvets2019leveraging}
Shvets, M., Liu, W., Berg, A.C.: Leveraging long-range temporal relationships
  between proposals for video object detection. In: ICCV. pp. 9756--9764 (2019)

\bibitem{simonyan2014two}
Simonyan, K., Zisserman, A.: Two-stream convolutional networks for action
  recognition in videos. In: Advances in neural information processing systems.
  pp. 568--576 (2014)

\bibitem{uijlings2013selective}
Uijlings, J.R., Van De~Sande, K.E., Gevers, T., Smeulders, A.W.: Selective
  search for object recognition. International Journal of Computer Vision
  \textbf{104}(2),  154--171 (2013)

\bibitem{vaswani2017attention}
Vaswani, A., Shazeer, N., Parmar, N., Uszkoreit, J., Jones, L., Gomez, A.N.,
  Kaiser, {\L}., Polosukhin, I.: Attention is all you need. In: NeurPIS. pp.
  5998--6008 (2017)

\bibitem{wang2018fully}
Wang, S., Zhou, Y., Yan, J., Deng, Z.: Fully motion-aware network for video
  object detection. In: ECCV. pp. 542--557 (2018)

\bibitem{wang2018non}
Wang, X., Girshick, R., Gupta, A., He, K.: Non-local neural networks. In: CVPR.
  pp. 7794--7803 (2018)

\bibitem{wu2019sequence}
Wu, H., Chen, Y., Wang, N., Zhang, Z.: Sequence level semantics aggregation for
  video object detection. In: ICCV. pp. 9217--9225 (2019)

\bibitem{xiao2018video}
Xiao, F., Jae~Lee, Y.: Video object detection with an aligned spatial-temporal
  memory. In: ECCV. pp. 485--501 (2018)

\bibitem{xu2015show}
Xu, K., Ba, J., Kiros, R., Cho, K., Courville, A., Salakhudinov, R., Zemel, R.,
  Bengio, Y.: Show, attend and tell: Neural image caption generation with
  visual attention. In: ICML. pp. 2048--2057 (2015)

\bibitem{zhu2018towards}
Zhu, X., Dai, J., Yuan, L., Wei, Y.: Towards high performance video object
  detection. In: CVPR. pp. 7210--7218 (2018)

\bibitem{zhu2017flow}
Zhu, X., Wang, Y., Dai, J., Yuan, L., Wei, Y.: Flow-guided feature aggregation
  for video object detection. In: ICCV. pp. 408--417 (2017)

\bibitem{zhu2017deep}
Zhu, X., Xiong, Y., Dai, J., Yuan, L., Wei, Y.: Deep feature flow for video
  recognition. In: CVPR. pp. 2349--2358 (2017)

\end{thebibliography}
\end{document}